\documentclass[letterpaper, 10 pt, conference]{ieeeconf}  

\IEEEoverridecommandlockouts                              

\pdfoutput=1

\newcommand{\Skip}[1]{}

\usepackage{tabularx}
\usepackage{graphicx}
\usepackage{amsmath}
\usepackage{amssymb}
\usepackage{booktabs}
\usepackage{bbding}
\usepackage{color,xcolor}
\usepackage{multirow}
\usepackage{float}
\usepackage{adjustbox}
\usepackage{url}
\usepackage{diagbox}
\usepackage[switch]{lineno}
\usepackage[pagebackref=true,breaklinks=true,colorlinks,bookmarks=false]{hyperref}
\usepackage[capitalize]{cleveref}

\crefname{section}{Sec.}{Secs.}
\Crefname{section}{Section}{Sections}
\Crefname{table}{Table}{Tables}
\crefname{table}{Tab.}{Tabs.}

\title{\LARGE \bf
MAexp: A Generic Platform for RL-based Multi-Agent Exploration 
}

\author{
Shaohao Zhu, Jiacheng Zhou, Anjun Chen, Mingming Bai, Jiming Chen, and Jinming Xu 
\thanks{$^{\dagger}$The authors are with the College of Control Science and Engineering, Zhejiang University, Hangzhou 310027, China. 
Correspondence to {\tt\small jimmyxu@zju.edu.cn} (Jinming Xu). This work was supported in part by NSFC under Grants 62088101, 62373323 and in
parts by the Key Laboratory of Collaborative Sensing and Autonomous
Unmanned Systems (Key Lab of CS\&AUS) of Zhejiang Province. Code and videos can be found at \href{https://github.com/DuangZhu/MAexp}{https://github.com/DuangZhu/MAexp}.
}%
}

\begin{document}

\newcommand{\reflabel}{dummy} 


\newcommand{\seclabel}[1]{\label{sec:\reflabel-#1}}
\newcommand{\secref}[2][\reflabel]{Section~\ref{sec:#1-#2}}
\newcommand{\Secref}[2][\reflabel]{Section~\ref{sec:#1-#2}}
\newcommand{\secrefs}[3][\reflabel]{Sections~\ref{sec:#1-#2} and~\ref{sec:#1-#3}}

\newcommand{\eqlabel}[1]{\label{eq:\reflabel-#1}}
\renewcommand{\eqref}[2][\reflabel]{(\ref{eq:#1-#2})}
\newcommand{\Eqref}[2][\reflabel]{(\ref{eq:#1-#2})}
\newcommand{\eqrefs}[3][\reflabel]{(\ref{eq:#1-#2}) and~(\ref{eq:#1-#3})}

\newcommand{\figlabel}[2][\reflabel]{\label{fig:#1-#2}}
\newcommand{\figref}[2][\reflabel]{Fig.~\ref{fig:#1-#2}}
\newcommand{\Figref}[2][\reflabel]{Fig.~\ref{fig:#1-#2}}
\newcommand{\figsref}[3][\reflabel]{Figs.~\ref{fig:#1-#2} and~\ref{fig:#1-#3}}
\newcommand{\Figsref}[3][\reflabel]{Figs.~\ref{fig:#1-#2} and~\ref{fig:#1-#3}}

\newcommand{\tablelabel}[2][\reflabel]{\label{table:#1-#2}}
\newcommand{\tableref}[2][\reflabel]{Table~\ref{table:#1-#2}}
\newcommand{\Tableref}[2][\reflabel]{Table~\ref{table:#1-#2}}
\newcommand{\etal}{et al.}
\newcommand{\eg}{e.g.}
\newcommand{\ie}{i.e. }
\newcommand{\etc}{etc. }

\def\bfmu{\mbox{\boldmath$\mu$}}
\def\bftau{\mbox{\boldmath$\tau$}}
\def\bftheta{\mbox{\boldmath$\theta$}}
\def\bfdelta{\mbox{\boldmath$\delta$}}
\def\bfphi{\mbox{\boldmath$\phi$}}
\def\bfpsi{\mbox{\boldmath$\psi$}}
\def\bfeta{\mbox{\boldmath$\eta$}}
\def\bfnabla{\mbox{\boldmath$\nabla$}}
\def\bfGamma{\mbox{\boldmath$\Gamma$}}

%
%


\newcommand{\R}{\mathbb{R}}

\maketitle
\thispagestyle{empty}
\pagestyle{empty}

\begin{abstract}
The sim-to-real gap poses a significant challenge in RL-based multi-agent exploration due to scene quantization and action discretization. Existing platforms suffer from the inefficiency in sampling and the lack of diversity in Multi-Agent Reinforcement Learning (MARL) algorithms across different scenarios, restraining their widespread applications. To fill these gaps, we propose MAexp, a generic platform for multi-agent exploration that integrates a broad range of state-of-the-art MARL algorithms and representative scenarios. Moreover, we employ point clouds to represent our exploration scenarios, leading to high-fidelity environment mapping and a sampling speed approximately 40 times faster than existing platforms. Furthermore, equipped with an attention-based Multi-Agent Target Generator and a Single-Agent Motion Planner, MAexp can work with arbitrary numbers of agents and accommodate various types of robots. Extensive experiments are conducted to establish the first benchmark featuring several high-performance MARL algorithms across typical scenarios for robots with continuous actions, which highlights the distinct strengths of each algorithm in different scenarios. 

\end{abstract}


\section{Introduction}

Multi-agent exploration is a rapidly growing field with various applications, such as search and rescue~\cite{liu2016multirobot} and environmental surveillance~\cite{fascista2022toward}. Despite its promising perspective, developing coordination policies that perform efficiently across diverse scenarios remains a challenge. Traditional methods, such as Potential Field-Based~\cite{yu2021smmr} and Cost-Based Exploration~\cite{mei2006energy,osswald2016speeding}, commonly suffer from the limited representational capacity for coordination policies. Their efficiency can be sensitive to specific conditions~\cite{tan2022deep} or require tedious parameter tuning~\cite{yu2022learning} tailored for individual scenarios, highlighting the requirement for more robust approaches.

Recent advancements in Multi-Agent Reinforcement Learning~\cite{samvelyan2019starcraft, baker2019emergent} have opened up avenues for improving the performance of multi-agent exploration. MARL algorithms have demonstrated their superior exploration efficiency over traditional methods in specific scenarios, such as indoor visual navigation~\cite{wang2021collaborative, yu2022learning} and exploration in discrete grid environments~\cite{mete2023coordinated, tan2022deep, chen2019end}.
However, these methods still exhibit a significant sim-to-real gap due to various factors, such as action discretization~\cite{wang2021collaborative}, scene quantization~\cite{mete2023coordinated, tan2022deep, chen2019end}, and the neglect of physical collision constraints~\cite{yu2022learning}.

The simulation scenarios utilized for MARL training and execution play a key role in bridging the sim-to-real gap~\cite{xu2022explore}. Different scenarios can provide diverse experiences, which is of great importance for robust strategies. On the other hand, the discrepancy between simulation scenarios and actual conditions directly influences the performance of derived strategies applied in the real world. However, most of the existing works rely heavily on non-standardized and randomly generated grid maps~\cite{zhang2022h2gnn,tan2022deep,mete2023coordinated}, which not only widens the sim-to-real gap but also limits the reproducibility and leads to unfair algorithmic comparisons. Other works utilizing open-source platforms with standardized maps for training and policy evaluation further encounter computational bottlenecks~\cite{xu2022explore,yu2022learning}, resulting in extensive training time, even with powerful hardware. Due to the tedious training process of MARL, the performance of various MARL algorithms in different exploration scenarios remains unexplored. This is an important issue to be solved to enable fair comparison among algorithms.

In this paper, we propose the first generic highly efficient \underline{M}ulti-\underline{A}gent \underline{exp}loration platform (MAexp) that integrates a variety of MARL algorithms and scenarios. To narrow the sim-to-real gap, we leverage the benefits of point-cloud representations over traditional grid-based methods, dynamically adjusting point-cloud density to represent diverse exploration areas. This design allows for high-fidelity mapping in intricate regions while maintaining computational efficiency with less complex landscapes. To allow our agent framework to handle teams of arbitrary size, and to train exploration policies for any type of robot, we formulate our exploration problem into a two-step procedure: i) generating navigation goals for agents using a global attention-based target generator; ii) calculating a tailored navigation path to the goal via a local motion planner compatible with any navigation algorithms.

Our primary contributions are three-folds:
\begin{itemize}
  \item We propose a generic high-efficiency platform for RL-based multi-agent exploration, accommodating various algorithms and scenarios, and achieving a sampling speed nearly 40 times faster than existing platforms.
  \item We employ an agent framework within the MAexp platform that can adapt to arbitrary team sizes and robot types during training.
  \item We establish a benchmark featuring six SOTA MARL algorithms and six typical scenarios,  setting a foundational standard for rigorous evaluation and comparison of multi-agent exploration techniques.
\end{itemize}
\section{Related Works}
\subsection{Reinforcement Learning in Multi-Agent Exploration} 
The existing frameworks for MARL in multi-agent exploration can be divided into two categories: Centralized Training with Centralized Execution (CTCE) \cite{chen2019end,geng2019learning} and Centralized Training with Decentralized Execution (CTDE) \cite{tan2022deep,yu2022learning,zhang2022h2gnn,he2020decentralized}. CTCE models, such as \cite{geng2019learning}, employ attention mechanisms in MARL for grid-based exploration. \cite{chen2019end} introduces CMAPPO, a CNN-augmented Proximal Policy Optimization variant, outperforming traditional frontier-based methods. However, CTCE is vulnerable to single-point failures and scalability issues because of its centralized architecture. CTDE methods are thus employed to address these limitations. For instance, \cite{he2020decentralized} proposes a novel distributed exploration algorithm using Multi-Agent Deep Deterministic Policy Gradients (MADDPG) \cite{lowe2017multi} for context-aware decision-making in structured settings. Subsequent studies, such as \cite{yu2022learning}, enhance this by incorporating attention mechanisms for spatial and inter-agent information processing, while \cite{tan2022deep} extends CTDE frameworks to tackle decentralized exploration in complex terrains.

However, the current research usually narrows its focus to individual MARL algorithms and specific exploration scenarios, thereby limiting both cross-algorithmic and cross-scenario evaluation. Conversely, our study leverages six SOTA MARL algorithms across six different exploration scenarios to establish the first comprehensive benchmark,  filling the notable gap in MARL exploration.

\subsection{Existing Platforms for Multi-Agent Exploration}
In the field of reinforcement learning based multi-agent exploration, platforms generally fall into two categories: vision-based exploration platforms using Habitat \cite{szot2021habitat}, as employed by MAANS \cite{yu2022learning}, and grid-based platforms with discrete maps \cite{tan2022deep,chen2019end,xu2022explore,zhang2022h2gnn}. While Habitat offers realistic simulations, it also involves additional modules, such as the SLAM module, which increases the sampling time during policy training. On the other hand, grid-based platforms suffer from a significant sim-to-real gap due to the substantial difference between the simulation scenarios and the real world. Moreover, most existing platforms support only a single MARL algorithm~\cite{yu2022learning,xu2022explore}, making cross-algorithmic comparisons impossible. In contrast, our platform provides a range of standard continuous exploration scenarios while offering multiple MARL algorithms for proper comparisons. It utilizes high-fidelity point-cloud-based scenarios and continuous actions to narrow the sim-to-real gap, and enhances computational efficiency for faster sampling. Note that the agent framework in our platform, built on the foundation of MARLLib~\cite{hu2022marllib}, is integrated with a local planner to account for various types of robots.

\section{Problem Formulation}
\subsection{Task Setup}
\begin{figure}[ht]
    \centering
    \includegraphics[width=\linewidth,trim=0 0 0 0]{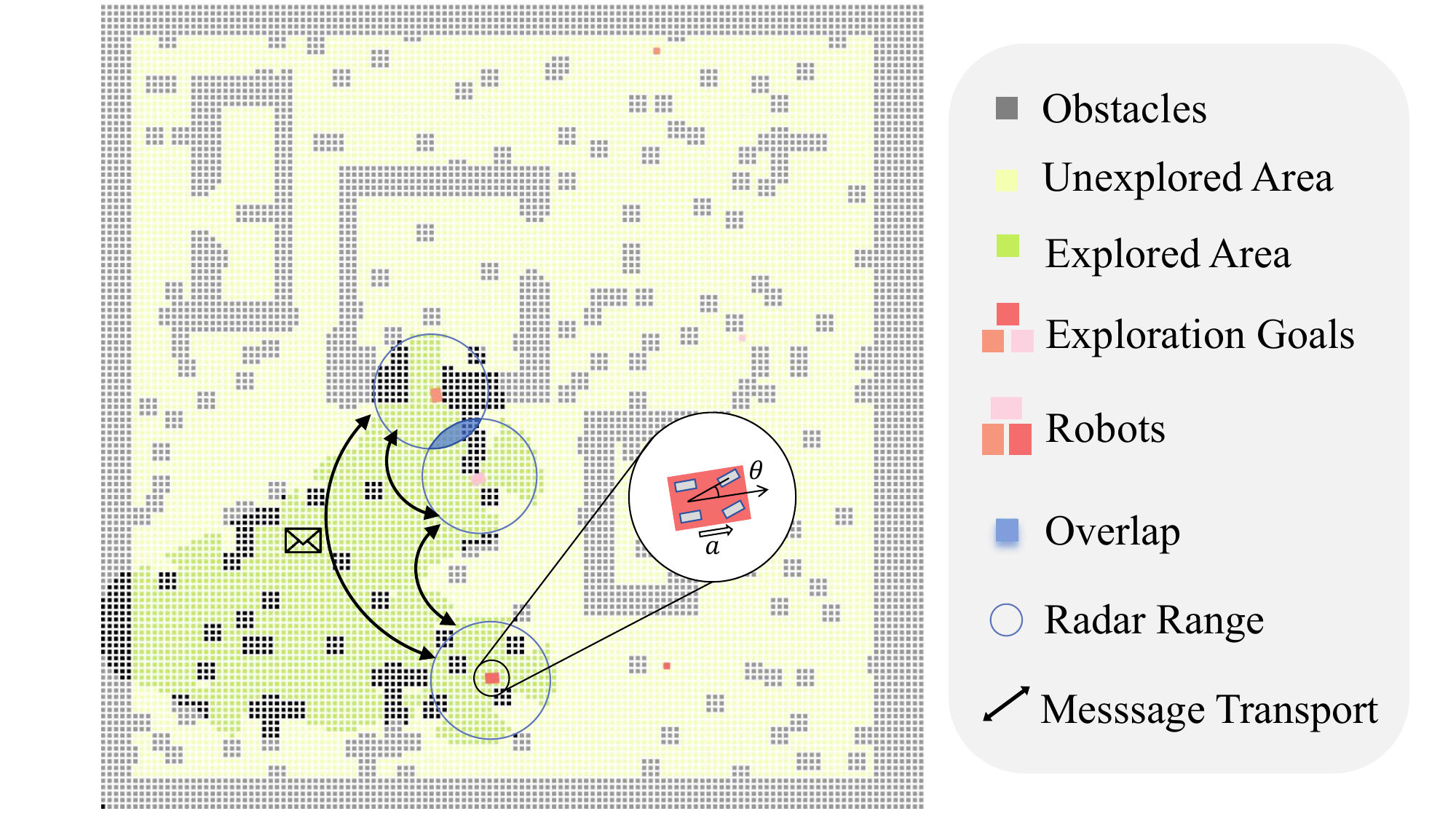}
    \vspace{-0.5cm} 
    \caption{Multi-agent coordination exploration in a random obstacle scenario.}
    \label{setup}
\end{figure}

We investigate a multi-agent coordination exploration problem, focusing on a team of robots tasked with cooperatively exploring an unknown continuous scenario as swiftly as possible, as illustrated in Fig. \ref{setup}. 
The explored areas are highlighted in green, with unexplored areas in yellow. Utilizing the Ackermann Car Model as a representative example, each agent performs two continuous actions: Acceleration and Steering Angle. At each time step, the agent acquires point-cloud data (shown as green points within radar range) from sensors like radars or cameras. In our framework, we assume perfect communication between agents, allowing for the exchange of observations, states, and goals to generate actions during both the training and execution phases. The primary objective of this task is to maximize the accumulated explored area (entire green region) while minimizing the overlap (indicated by the blue area) between agents within a constrained time horizon. 

\subsection{Mathematical Formulation}
While implementing MARL algorithms, it is essential to model the problem as a decentralized partially observable Markov decision process (Dec-POMDP). A Dec-POMDP is defined by the tuple \( \left \langle n, S, A, O, R, P, \gamma, h \right \rangle \). Here, \( n \) denotes the number of agents involved in the task, and \( S \) signifies the state space, with \( s_i \) representing the state of agent \( i \) and \( s \) representing the joint state of all agents. The joint action space is symbolized by \( A \), and the individual actions of agent \( i \) and joint action are represented by \( a_i \) and \( a \). An observation for agent \( i \), denoted by \( o_i \), is generated by the observation function \( O(s, a_i) \), dependent on both the states of all agents and their corresponding actions. The reward for agent \( i \) and the whole team, symbolized by \( r_i \) and \( r \), can be derived from the reward function \( R(s,a) \). The transition probability from state \( s \) to state \( s' \) through action \( a \) is defined by \( P(s' | s, a) \), dependent on the environment's properties. The discount factor for future rewards is represented by \( \gamma \), and \( h \) indicates the horizon of an episode. The goal of this problem is for each agent to find the most effective strategy, denoted as \( \pi_i \), and collaboratively work towards maximizing the collective reward for the team as follows:
\[
J(\theta) = \mathbb{E}\left[\sum_{t = 0}^{h-1} \gamma^t R(s(t),a(t)) \bigg| s(0),\pi\right].
\]
The ideal coordination strategy is formulated using gradient descent to optimize the objective function \( J(\theta) \) relative to \( \pi \).

\subsection{Metrics}
To evaluate exploration policies, we divide our metrics into two dimensions: Team-Based and Agent-Specific metrics. These metrics evaluate both the efficiency and cooperation abilities of the team within the exploration process.
\begin{figure*}
\centering
\setlength{\abovecaptionskip}{0.1mm}
\setlength{\belowcaptionskip}{-6mm}
    \includegraphics[trim={0.2cm 2.7cm 0.2cm 3.5cm},clip,width=0.9\linewidth]{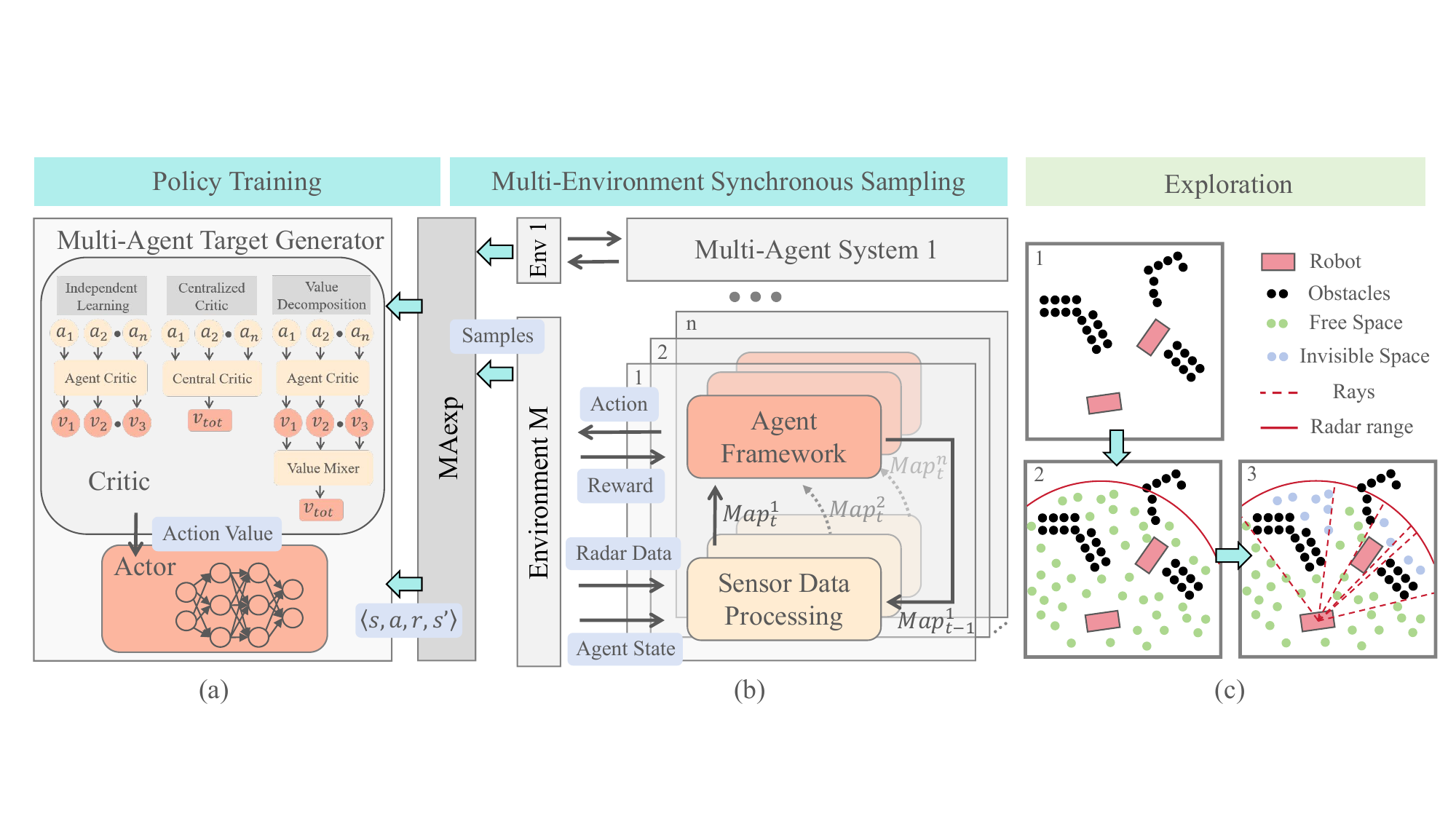}
    \caption{The proposed MAexp platform for multi-agent exploration.}
    \label{fig:framework}
\end{figure*}

\subsubsection{Team-Based Metrics}
\begin{itemize}
    \item \textbf{Exploration Ratio (ER):} Quantifies the proportion of the explored area to the total explorable space at the end of an episode.
    \item \textbf{Coverage Step (CS):} Denotes the number of steps required for the team to reach the predefined exploration thresholds (85\% and 95\%). An 85\% ratio implies significant topological coverage, while 95\% marks the episode's successful completion.
\end{itemize}

\subsubsection{Agent-Specific Metrics}
\begin{itemize}
    \item \textbf{Mutual Overlap (MO):} Measures the ratio of the area only explored by one agent to the total explored area. Less overlap suggests better task allocation and minimized redundancy. This metric is recorded when the exploration ratio reaches 85\% and 95\%.
    \item \textbf{Reward standard Variance (RV):} Captures the variability in rewards across agents at the end of an episode. A lower variance indicates a more balanced contribution from each agent, avoiding situations where certain agents underperform but are overshadowed by the team's collective results.
\end{itemize}

\subsection{Reward Function}
Our reward function uniquely fuses team performance with agent contributions, comprising five elements: Success Reward, Exploration Reward, Overlap Penalty, Collision Penalty, and Time Penalty. Let \(Cov_t\) represent the total coverage ratio at time \(t\), \(map^{tot}_t\) the merge map, and \(map^{i}_{t}\) the area explored by agent \(i\) at the same time step. In all \(map\) variables, '1' denotes explored space, while '0' indicates unexplored territory.

The agent-specific total reward \(R_{\text{total}}^{i}\) is formalized as:
\[
R_{\text{total}}^{i} = 
\begin{cases}
\begin{aligned}
& R_s, & & \text{(Success Reward)} \\
& \delta map^{i}_{t}, & & \text{(Exploration Reward)} \\
& -\sum(P^{-i}_{t} \cap P^{i}_{t}), & & \text{(Overlap Penalty)} \\
& -R_c,  & & \text{(Collision Penalty)} \\
& -Cov_t, & & \text{(Time Penalty)}
\end{aligned}
\end{cases}
\]

Here, \(\delta map^{i}_{t}\) is proportional to \(\sum (map^{i}_{t} - map^{tot}_{t-1})\), where the summation is over all locations where the difference equals 1. \(P^{i}_{t}\) and \(P^{-i}_{t}\) denote the point clouds gathered by agent \(i\) and the rest of the team at time \(t\), respectively. Note that the final agent-specific reward is a normalized linear combination of these five components.

\section{The Proposed Platform}
\subsection{The proposed MAexp Platform}
We introduce MAexp, a generic high-efficiency platform designed for multi-agent exploration, encompassing a diverse range of scenarios and MARL algorithms. The platform is developed in Python to smoothly integrate with existing reinforcement learning algorithms, and it is equally applicable to traditional exploration methods. In an effort to bridge the sim-to-real gap, all maps and agent properties within MAexp are modelled continuously, incorporating realistic physics to closely mirror real-world exploration.

The autonomous exploration simulation employs an integrated data flow, as shown in Fig. \ref{fig:framework} (a-b). Part (b) presents the platform's sampling mechanism. In this phase, MAexp initiates multiple environments to collect vast experience. Agents, at every time step, acquire point-cloud data and their respective states from the environment. Moreover, they can adjust the radar resolution and detection range based on specific situations. The point-cloud data generation during simulation is depicted in Fig. \ref{fig:framework} (c). This procedure simulates radar perception but operates at a faster speed. Once an agent completes an action, the environment identifies all free space point clouds within the detection range of the agent. Subsequent filtering eliminates points masked by obstacles using the following criteria:
\[
\begin{cases}
\frac{P_f \cdot P_o}{\|P_f\| \|P_o\|}  > 1-\alpha_1;\ \\
\| P_f \| - \| P_o \| <  \alpha_2,
\end{cases}
\]
where \(P_f\) denotes the vector from the robot to a free space point, \(P_o\) represents the vector to an obstacle, and $\alpha_1,\alpha_2$ are certain parameters to be determined. This process filters out obstructed points (c.f., blue points in Fig. \ref{fig:framework} (c)) and transmits the remaining ones to the agent. Utilizing past feature maps and environmental data, the current feature map is generated and distributed to all agents within the environment. The agent frameworks consequently produce the actions for agents. Implementing these actions provides individual rewards for every agent, mirroring the action value. Thereafter, the environment transitions the collective agents to the succeeding states. MAexp, through consistent sampling, MAexp accumulates a substantial set of quadruples \( \left \langle s,a,r,s' \right \rangle\) for training policies. Fig. \ref{fig:framework} (a) illustrates the training process and provides an in-depth examination of the Multi-Agent Target Generator in the agent framework. During training, batches of quadruples are transported to both the actor and the critic to facilitate parameter updating. The critic provides the actor with the action values, which are essential for the gradient descent step.

It's worth noting that the proposed agent framework is well-adapted to a range of robots and group sizes. In particular, we address the multi-agent exploration challenge leveraging the new techniques developed for robotic navigation \cite{sadek2023multi,nakhleh2023sacplanner,shu2019incrementally,zhang2019self}. Therefore, we divide our agent framework into two modules: \emph{Multi-Agent Target Generator (where-to-go)} and \emph{Single-Agent Motion Planner (how-to-go)}. In the first module, agents use MARL algorithms to generate their own navigation points. In the second module which is independent of MARL, agents decide how to arrive these target locations by determining their acceleration and steering angle. Notably, the how-to-go module can incorporate any robot navigation algorithm for motion planning.


\subsection{Scenarios in MAexp}

\begin{figure}[ht]
    \centering
    \hspace*{-6mm}
    \includegraphics[width=1.2\linewidth,trim=0 0 0 0,clip]{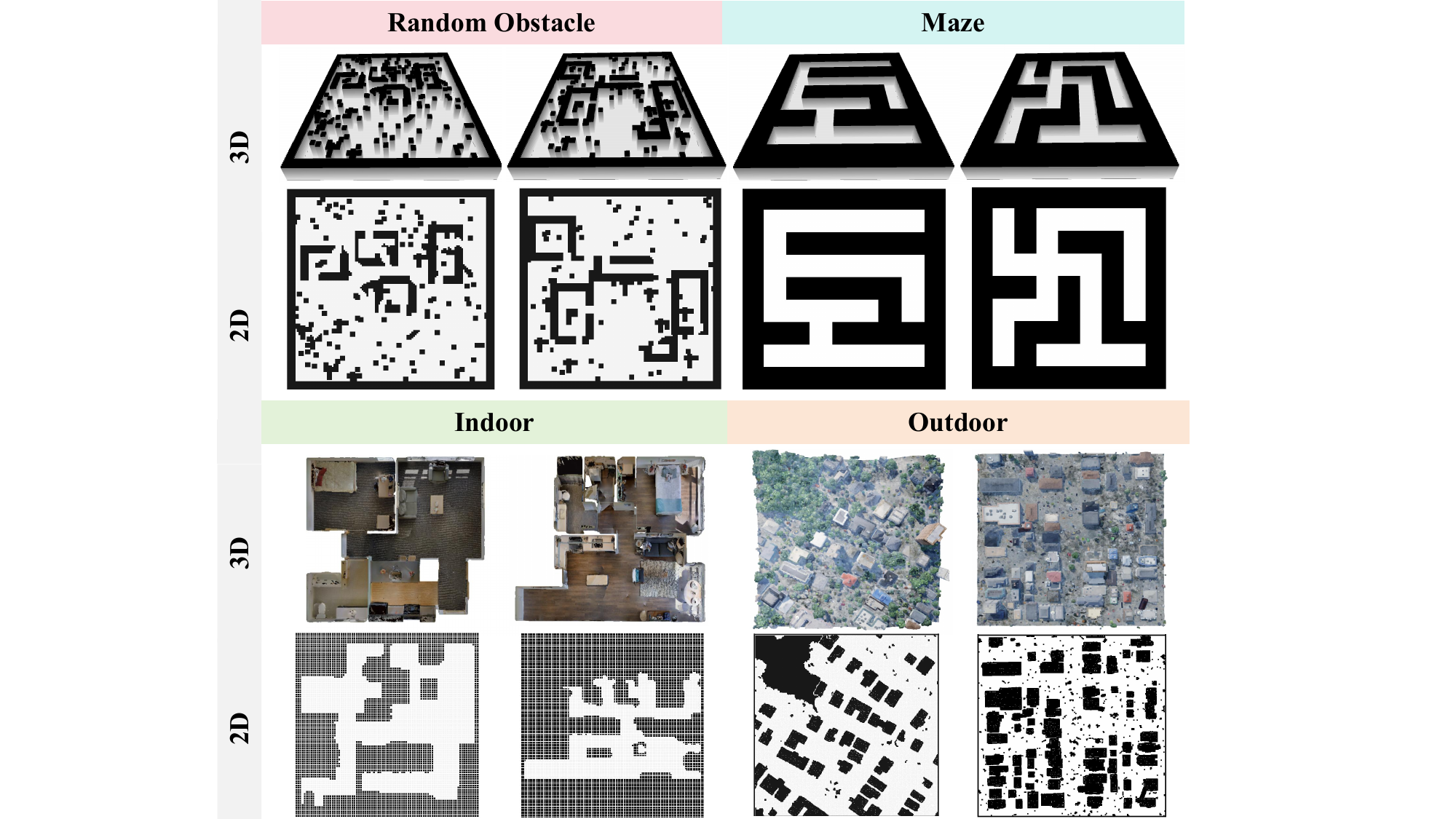}
    \caption{Illustration of exploration scenarios in MAexp with 3D visualizations and 2D top-down projection maps.}
    \label{scene}
\end{figure}

Different from previous studies that rely on grid-based scenarios, we utilize point clouds for all map formulations. This method ensures a seamless depiction of exploration areas, delivering both detailed and authentic representations. The processing of point clouds is easily aligned with GPU parallel processing, and the map's sparsity does not require global consistency. In obstacle-dense regions, for instance, there's no need to detail every obstacle—defining boundaries suffices. In pivotal regions, increasing the density of the point cloud can bolster map authenticity. Furthermore, point-cloud maps can be effortlessly extended into three-dimensional spaces. The detailed characteristics of the four unique exploration scenarios in MAexp, along with their 3/2D maps, can be found in TABLE \ref{scenes} and Fig. \ref{scene}.
\begin{table}[htbp]
  \centering
  \caption{Summary of various multi-agent exploration scenarios.}
    \begin{tabular}{cc|cccc}
    \toprule
    \multicolumn{2}{c|}{Properties} & Type  & Resolution & Size(m) & Quantity \\
    \midrule
    \multicolumn{2}{c|}{Random Obstacle} & SPC   & 1.0   & 125   & 32 \\
    \multicolumn{2}{c|}{Maze} & SPC   & 1.0   & 125   & 80 \\
    \multicolumn{1}{c|}{\multirow{3}[0]{*}{Indoor}} & Small & RPC   & 0.1   & \textless10   & 17 \\
    \multicolumn{1}{c|}{} & Medium & RPC   & 0.1   & 10-14 & 32 \\
    \multicolumn{1}{c|}{} & Large & RPC   & 0.1   & \textgreater14   & 22 \\
    \multicolumn{2}{c|}{Outdoor} & SPC \& RPC   & 1.5-2 & 288  & 32 \\
    \bottomrule
    \end{tabular}%
\\ 
\vspace{1mm}
*RPC: Real-world Point Clouds; SPC: Synthetic Point Clouds.
  \label{scenes}%
\end{table}%

\textbf{Random Obstacle scenarios} combine basic elements like narrow corridors, corner loops, and multiple rooms, as detailed in \cite{xu2022explore}. Then we intersperse isolated obstacles until the map achieves the desired obstacle density. This scenario is specifically designed to encapsulate a diverse range of exploration difficulties commonly faced by agents. 

\textbf{Maze scenarios} are generated through a refined Kruskal's algorithm. With numerous branching paths and the absence of closed loops, the Maze necessitate agents to strategically allocate different agents to distinct paths to improve exploration efficiency.

\textbf{Indoor scenarios} are derived from Habitat \cite{szot2021habitat}, a high-fidelity dataset of 3D indoor scans.
By closely mirroring actual indoor settings, these scenarios challenge the agent to conduct thorough exploration in constrained areas. Based on the map size, we categorize these scenarios into three types: large, medium, and small. 

\textbf{Outdoor scenarios} are derived from STPLS3D \cite{Chen_2022_BMVC}, which comprises a comprehensive collection of synthetic and actual aerial photogrammetry 3D point clouds. These maps authentically emulate extensive outdoor exploration settings, thereby challenging the system's adeptness at swift, large-scale outdoor exploration. 

It should be noted that during the process of map creation, we conducted careful checks and repairs to ensure the maps were suitable for exploration tasks, which is time-consuming but essential in achieving accurate results.

\begin{figure*}[!htbp]
\centering
\setlength{\abovecaptionskip}{0.1mm}
\setlength{\belowcaptionskip}{-8mm}
    \includegraphics[trim={0.5cm 5.5cm 0.5cm 0.8cm},clip,width=0.9\linewidth]{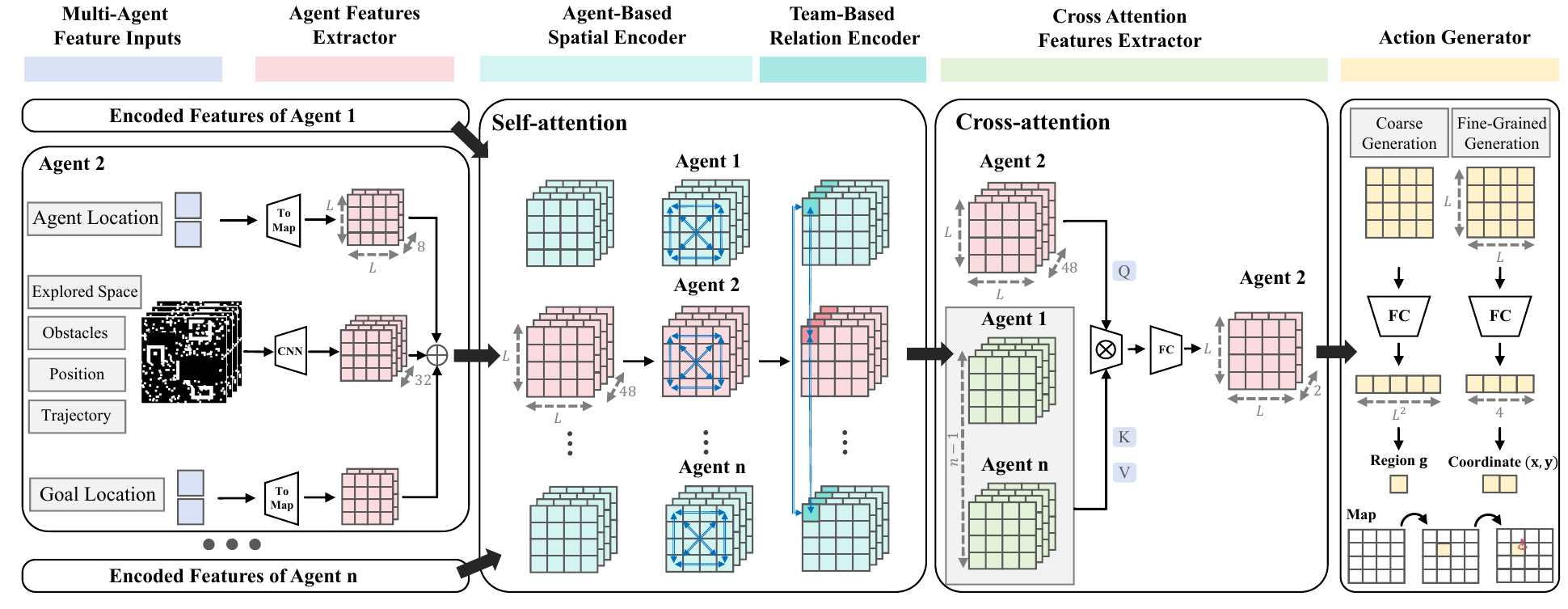}
    \caption{The overall structure of Multi-agent Target Generator.}
    \label{fig:att framework}
\end{figure*}

\subsection{MARL Algorithms in MAexp}

MARL algorithms can be broadly classified into three categories based on their critic architectures: Independent Learning (IL), Centralized Critic (CC), and Value Decomposition (VD). These architectures are illustrated in Fig. \ref{fig:framework}(a).

\textbf{Independent Learning} such as IPPO \cite{schulman2017proximal}, IA2C \cite{mnih2016asynchronous}, and ITRPO \cite{schulman2015trust}, operate with agent-level critic, focusing on singular observations. IPPO, for instance, exhibits robustness to certain environmental variability \cite{de2020independent}, while ITRPO dominates in the MAMuJoCo \cite{peng2021facmac} benchmark for multi-agent robotic control \cite{hu2022marllib}.

\textbf{Centralized Critic}, including MAPPO \cite{yu2022surprising} and MATRPO \cite{kuba2021trust}, employ a unified critic that fuses information from the entire team to judge the joint action value. Generally, CC algorithms outperform IL in coordinated tasks \cite{hu2022marllib,yu2022surprising}.

\textbf{Value Decomposition}, including VDPPO \cite{ma2022value} and VDA2C \cite{su2021value}, combine outputs from agent-level critics to estimate a team action value, thereby balancing both individual and group objectives. Despite their best performance in numerous benchmarks \cite{samvelyan2019starcraft,lowe2017multi,hu2022marllib}, they struggle with continuous control and long-term planning, typically seen in multi-agent exploration \cite{hu2022marllib}.

Recognizing the unique strengths of individual algorithms, we have incorporated into the proposed MAexp platform six leading MARL algorithms, such as IPPO, ITRPO, MAPPO, MATRPO, VDPPO, and VDA2C.

\subsection{Multi-Agent Target Generator}
The \emph{Multi-agent Target Generator}, whose entire structure is illustrated in Fig. \ref{fig:att framework}, employs a CNN model to extract 125$\times$125 grid-based spatial features from each agent, with inputs capturing explored space, obstacles, agent position, and trajectory. These feature maps, when combined with embedded agent and previous goal location, yield an $L \times L \times D$ feature map where \(D = 48\). Subsequently, team-wise data is fused via a hierarchical transformer-based structure, similar to \textit{Spatial-TeamFormer} in MAANS \cite{yu2022learning}, known for its superior performance in visual exploration. The \textit{Spatial-Teamformer block} integrates two layers: the \emph{Individual Spatial Encoder}, which applies spatial self-attention to each agent's $L \times L \times D$ map without inter-agent computations, and the \emph{Team Relation Encoder}, targeting team interactions without spatial considerations. These team-oriented data are then merged for each agent through cross-attention. In our approach, the transformer depth is fixed at 2, yielding a $2 \times L \times L$ grid feature as the output.

The \(2 \times L \times L\) grid feature feeds a dual-action generator to produce an exploration target. Using its first channel, the agent determines a discrete region for coarse generation. This feature translates to an \(L^2\) vector, denoting the probability of each region's selection, from which a region is sampled. Subsequently, for fine-grained generation, a coordinate \( (x, y) \) denotes the global goal's relative position within the chosen region. This coordinate is derived from a 4-vector, where the initial pair indicates the mean and variance for coordinate \(x\), and the latter pair for \(y\). Note that the above two generators operate in parallel and thus it is not necessary to feed the coarse action into the fine-grained generator.
\begin{table*}[!htbp]
\setlength{\abovecaptionskip}{0.1mm}
  \centering
    \caption{Performance results of six MARL algorithms across six scenarios. Within each scenario, the upper three columns represent `ER' (↑), `85\% CS' (↓), and `95\% CS' (↓), while the lower columns correspond to `85\% MO' (↓), `95\% MO' (↓), and `RV' (↓). }
  \resizebox{\textwidth}{!}{
    \huge
    \begin{tabular}{c|c|ccc|ccc|ccc|ccc|ccc|ccc}
    \toprule
    \multicolumn{2}{c|}{\multirow{2}[2]{*}{Scenarios}} & \multicolumn{6}{c|}{Generated Scenes}             & \multicolumn{12}{c}{Real Scenes}  \\
    \multicolumn{2}{c|}{} & \multicolumn{3}{c}{ \textbf{Random Obstacles}} & \multicolumn{3}{c|}{ \textbf{Maze}} & \multicolumn{3}{c}{ \textbf{Indoor-Small}} & \multicolumn{3}{c}{ \textbf{Indoor-Medium}} & \multicolumn{3}{c}{ \textbf{Indoor-Large}} & \multicolumn{3}{c}{ \textbf{Outdoor}}  \\
    
    \multicolumn{2}{c|}{} & \multicolumn{1}{c}{ER~(\%)} & \multicolumn{1}{c}{85\%CS~(step)}& \multicolumn{1}{c}{95\%CS~(step)}& \multicolumn{1}{c}{ER~(\%)} & \multicolumn{1}{c}{85\%CS~(step)}& \multicolumn{1}{c|}{95\%CS~(step)}& \multicolumn{1}{c}{ER~(\%)} & \multicolumn{1}{c}{85\%CS~(step)}& \multicolumn{1}{c}{95\%CS~(step)}& \multicolumn{1}{c}{ER~(\%)} & \multicolumn{1}{c}{85\%CS~(step)}& \multicolumn{1}{c}{95\%CS~(step)}& \multicolumn{1}{c}{ER~(\%)} & \multicolumn{1}{c}{85\%CS~(step)}& \multicolumn{1}{c}{95\%CS~(step)}& \multicolumn{1}{c}{ER~(\%)} & \multicolumn{1}{c}{85\%CS~(step)}& \multicolumn{1}{c}{95\%CS~(step)}  \\
    \midrule
    
    \multirow{4.5}[4]{*}{Team-Based} & ITRPO &$\textbf{68.81}\pm5.71$ & -  & -  & $79.71\pm7.32$ &$	449\pm63 $&$	501\pm77 $& $90.46\pm5.12$   & $268\pm45$  & $437\pm65$  & $79.38\pm5.00$  &$400\pm84$   & $557\pm47$   &$59.40\pm3.95$  & -  & -   & $26.91\pm4.12$ &-  &- \\
    
    & IPPO &$63.07\pm4.69$   & -  & -    & $\textbf{93.17}\pm4.27$	& $377\pm51 $& $441\pm57$
   &$\textbf{94.18}\pm0.76$   & $\textbf{157}\pm5$  & $\textbf{205}\pm41$  & $\textbf{91.48}\pm4.36$  &$\textbf{287}\pm54$   & $470\pm61$   &$71.32\pm4.79$  & -  & -   & $33.06\pm6.90$ &-  &- \\
   
    & MATRPO& $61.12\pm3.76$   & -  & -   & $86.29\pm7.63$	&$376\pm 72$&$	472\pm65 $ &$89.66\pm0.79$   & $237\pm13$  & $356\pm92$  & $75.73\pm8.91$  & $472\pm44$   & $536\pm67$   & $\textbf{77.31}\pm4.39 $ & -  & -   &$32.72\pm4.31$ &-  &-\\
    
    & MAPPO & $62.53\pm3.38$   & -  & -   & $89.15\pm5.07$ &	$370\pm67$ & $442\pm44$ & $93.23\pm3.56$   &$206\pm88$  &$284\pm107$  & $87.01\pm5.71$  & $356\pm40$   & -   & $70.19\pm4.94$  & -  & -   & $\textbf{38.15}\pm7.24$ &-  &-\\
   
    & VDPPO &$60.04\pm4.54$ & - & -& $83.12\pm5.08$   & $468\pm45$  & $506\pm35$   & $91.85\pm3.01$	& $236\pm92$ & $358\pm111$
   & $89.97\pm3.11$   & $308\pm41$  & $\textbf{447}\pm24$  & $67.72\pm4.96$  &-   & -      & $30.31\pm5.48$ &-  &-\\
   
     & VDA2C & $64.23\pm3.43$   & -  & -   &$\textbf{93.23}\pm3.54$	& $\textbf{322}\pm59$ & $\textbf{416}\pm54$
   &$\textbf{93.59}\pm0.33$   & $170\pm12$  & $235\pm52$  & $76.93\pm3.95$  & $320\pm30$   &  $\textbf{449}\pm26$   & $73.97\pm2.86$  & -  & -   & $26.76\pm4.38$ &-  &-\\
    \midrule
    \multicolumn{2}{c|}{} & \multicolumn{1}{c}{85\%MO~(\%)} & \multicolumn{1}{c}{95\%MO~(\%)}& \multicolumn{1}{c}{RV}& \multicolumn{1}{c}{85\%MO~(\%)} & \multicolumn{1}{c}{95\%MO~(\%)}& \multicolumn{1}{c|}{RV}& \multicolumn{1}{c}{85\%MO~(\%)} & \multicolumn{1}{c}{95\%MO~(\%)}& \multicolumn{1}{c}{RV}& \multicolumn{1}{c}{85\%MO~(\%)} & \multicolumn{1}{c}{95\%MO~(\%)}& \multicolumn{1}{c}{RV}& \multicolumn{1}{c}{85\%MO~(\%)} & \multicolumn{1}{c}{95\%MO~(\%)}& \multicolumn{1}{c}{RV}& \multicolumn{1}{c}{85\%MO~(\%)} & \multicolumn{1}{c}{95\%MO~(\%)}& \multicolumn{1}{c}{RV}  \\    
    \midrule
    \multirow{4.5}[4]{*}{Agent-Specific} &ITRPO &-   & -  & $\textbf{610.73}\pm79.25$   &$51.56\pm6.49$ & $43.89\pm4.76$	&$931.00\pm104.18$
    &$\textbf{51.37}\pm4.84$   & $\textbf{48.90}\pm2.54$  & $871.78\pm38.41$  & $49.60\pm7.49$  & $44.77\pm6.84$   & $347.76\pm42.10$  & -  & -  & $252.95\pm104.43$   & - &-  &$457.93\pm84.23$ \\
    & IPPO& -   & -  & $711.56\pm109.97$  & $\textbf{45.52}\pm7.87$ &	$\textbf{42.38}\pm7.12$ & $970.43\pm76.31$
    &$54.06\pm3.55$   & $50.89\pm4.59$  & $576.31\pm8.84$  & $46.38\pm3.35$  & $46.39\pm4.36$   & $318.29\pm50.87$   & -  & -  & $257.30\pm59.91$   & - &-  &$534.92\pm94.25$ \\
   
    & MATRPO & -   & -  & $825.65\pm63.06$   & $48.21\pm9.19$ &	$43.91\pm8.12$ &	$910.37\pm82.33$ & $57.49\pm4.59$   & $59.37\pm4.29$  & $547.39\pm45.81$  & $49.92\pm8.45$  & $57.92\pm8.39$   & $389.70\pm89.74$   & -  & -  & $280.52\pm74.28$   & - & -  &$433.80\pm115.81$\\
    
    & MAPPO & -   & -  & $831.93\pm94.97$   & $46.22\pm9.46$ &	$38.53\pm2.95$ & $929.25\pm89.14$
   &$60.78\pm9.24$   & $59.14\pm8.47$  & $602.14\pm78.95$  & $38.22\pm2.81$  & -   & $385.22\pm49.97$   & -  & -  & $248.80\pm87.86$   & - &-  &$445.98\pm118.32$ \\
          & VDPPO & -   & -  & $\textbf{632.18}\pm83.41$   & $58.96\pm5.03$	& $58.66\pm4.02$ & $951.11\pm82.10$
   &$64.03\pm15.29$   & $66.83\pm17.67$  & $541.55\pm99.39$  & $\textbf{36.58}\pm1.63$  & $42.38\pm0.74$ & $\textbf{309.89}\pm31.52$   & -  & -  & $\textbf{198.38}\pm78.41$   & - &-  &$\textbf{431.67}\pm103.54$ \\
          & VDA2C & -   & -  & $810.62\pm76.26$   & $49.60\pm5.65$	& $47.62\pm5.91$ & $\textbf{907.74}\pm60.99$
    & $77.14\pm3.89$  & $77.32\pm8.40$   & $\textbf{496.54}\pm33.77$ & $45.20\pm3.32$   & $\textbf{40.45}\pm1.04$  & $445.24\pm39.86$   & -  & -  & $275.55\pm37.27$   & - &-  & $561.50\pm114.43$ \\

    \bottomrule
    \end{tabular}}%
  \label{main result}%
\end{table*}%

\section{Experiments}
\subsection{Environmental Setup}
In our experimental studies, we employ a swarm of Ackermann cars equipped with adjustable radars, tailoring the resolution and detection range to various exploration scenarios. The \emph{Multi-Agent Target Generator} operates as described earlier, while the \emph{Single-Agent Motion Planner} employs the Dynamic Window Approach (DWA), which provides precise navigation and obstacle avoidance.

Our experiments involve \(N = 3\) agents, with parameters set at \(L = 8\), \(\alpha_1 = 10^{-3}\), \(\alpha_2 = 10^{-6}\), \(R_s = 100\), and \(R_c = 200\). Each MARL training spans \(10^4\) iterations across three random seeds. Results are presented as ``mean (standard deviation)," averaged from 300 tests—100 per seed. We apply all six MARL algorithms across 16 settings: 4 Random Obstacle, 4 Maze, 3 Outdoor, and 5 Indoor. We further classify the ``Indoor'' into 2 small, 2 medium, and 1 large. Training is carried out on an Ubuntu 18.04 server with two NVIDIA GeForce RTX 3090 GPUs: one for multi-environment sampling and another for online training. Each exploration map receives specialized parameter training to optimize algorithm performance. A \(10^4\) iteration training requires about 60 hours, culminating in a total of roughly 750 days of continuous GPU usage across over 300 runs.

\subsection{Comparison of Simulation Speed}
We first compare the sampling speeds of several open-source SOTA platforms for MARL exploration, i.e., MAexp, MAANS \cite{yu2022learning}, and Explore-Bench \cite{xu2022explore}.

TABLE. \ref{simulation comparison} illustrates the sampling times per step for team sizes \(N = 2, 4, 6, 8\). The proposed MAexp platform, optimized for MARL exploration, achieves a simulation speed nearly 40 times faster than MAANS. While both utilize the same exploration principles, MAANS, designed for vision tasks, incur additional computational costs. In contrast, Explore-Bench's level-0 component, similar to ours, is tailored for MARL sampling. Unlike Explore-Bench's CPU-centric grid simulations, MAexp leverages point-cloud modelling and GPU parallelization, substantially accelerating simulations. Hence, MAexp emerges as the most efficient platform for MARL exploration, facilitating the evaluation and development of new MARL algorithms. Note that our platform can also accommodate a large number of robots as long as  communication and action generation strategies are properly adjusted for enhanced efficiency.

\begin{table}[!htpb]
\centering
\caption{Comparison of Sampling times(s) for different platforms}
\begin{tabular}{lcccc}
\toprule
                  & \( N=2 \)       & \( N=4 \)       & \( N=6 \)      & \( N=8 \)       \\
\midrule
\textbf{MAANS~\cite{yu2022learning}}     & 0.4497     & 0.9319     & 1.3675     & 1.9354     \\
\textbf{Explore-Bench~\cite{xu2022explore}} & 3.8586           & 6.6381             & 8.6575             & 10.9880              \\
\textbf{MAexp (ours)}     & \textbf{0.0144}             & \textbf{0.0254}             & \textbf{0.0373}             & \textbf{0.0477}              \\
\bottomrule
\end{tabular}

\label{simulation comparison}
\end{table}

\subsection{Performance of MARL Algorithms in Various Scenarios}
In Table \ref{main result}, we observe that different algorithms exhibit different characteristics across our proposed scenarios. In particular, ITRPO performs impressively well in the ``Random Obstacles'' scenario, while IPPO consistently achieves the best exploration ratios in ``Maze'', ``Indoor-Small,'' and ``Indoor-Medium ''. These IL-based approaches excel in small-scale exploration scenarios characterized by dense challenges such as corner loops and multiple rooms, as their critics primarily focus on the agent's immediate environment to generate suitable strategies.

VD-based algorithms, such as VDPPO and VDA2C, tend to obtain efficient exploration policies by uniform task distribution, especially when agents maintain consistent performance. This is evident in the consistently low ``Reward Variance'' observed in VDPPO and VDA2C across all scenarios, even though they exhibit lower exploration ratios. Further, the superior performance of VDA2C in ``Maze'' reinforces the perspective since the efficiency of exploration among agents becomes uniformly distributed in this structured environment with a constant width of free space. 

In contrast, CC-based approaches, such as MATRPO and MAPPO, demonstrate superior performance in large-scale scenarios characterized by sparsely distributed obstacles as they tend to disperse agents to different regions of the map for parallel exploration. As evidenced in the Table, MATRPO achieves a 77.31\% exploration ratio in ``Indoor-Large'' while MAPPO excels in ``Outdoor''. These algorithms allow for an overall understanding of the situation of the entire team by fusing the observations and states among agents, facilitating efficient spatial allocation. However, the fusion also increases the complexities of policy training, making it challenging to adopt strategies tailored to immediate surroundings, resulting in lower performance in small-scale scenarios.

To sum up, understanding the characteristics of the exploration scenario is crucial for choosing a proper MARL algorithm to generate robust and efficient coordination policies. Once the scenario is identified, the proposed MAexp platform offers a valuable tool for evaluating candidate MARL algorithms comprehensively, facilitating the selection of the most appropriate one. Moreover, for those involved in designing new algorithms, MAexp also serves as a generic, high-efficiency platform for both training and simulation, as well as a benchmark for performance comparison.


\section{CONCLUSIONS}
We introduced MAexp, a generic high-efficiency platform for multi-agent exploration. MAexp incorporates several state-of-the-art MARL algorithms and various representative exploration scenarios, and it employs point-cloud representation for maps which enhances the effectiveness of MARL algorithms with rapid sampling and realistic simulation environments. Moreover, with its well-designed agent framework, MAexp can accommodate a variety of robots and group sizes. Furthermore, we establish the first comprehensive benchmark featuring several high-performance MARL algorithms across various typical scenarios. Our results highlight the unique strengths of each algorithm in different scenarios. In our future work, we aim to enhance MAexp to account for general communication topology and incorporate more advanced MARL algorithms and practical scenarios so as to provide a versatile platform for multi-agent exploration. We believe that our platform can advance the field of RL-based multi-agent exploration.





\bibliographystyle{IEEEtran}
\bibliography{ref}


\end{document}